\documentclass[letterpaper, 10 pt, conference]{ieeeconf}  % Comment this line out if you need a4paper
\IEEEoverridecommandlockouts
\usepackage[utf8]{inputenc} % usually not needed (loaded by default)
\usepackage[T1]{fontenc}
\usepackage{amsmath}
\usepackage{amsfonts}
\usepackage{float}
\usepackage{graphicx}
\usepackage{booktabs}
\usepackage[font=small]{caption}
\usepackage{subcaption}
\usepackage[linesnumbered, ruled, vlined]{algorithm2e}
\usepackage{booktabs} 
\usepackage{enumerate}
\usepackage{times}
\usepackage{soul}
\usepackage{url}
\usepackage{hyperref}
\usepackage{bm}
\usepackage{adjustbox}
\usepackage{xcolor}
\usepackage{placeins}
\usepackage[utf8]{inputenc}
\usepackage{booktabs}
\usepackage{array, multirow}
\usepackage{cleveref}
\usepackage{tikz}
\usepackage{wrapfig}
\usepackage{adjustbox}
\usepackage{multirow}
\usepackage{amssymb}% http://ctan.org/pkg/amssymb
\usepackage{pifont}
\usepackage{adjustbox}

\usepackage{caption}

\newcommand{\bb}{\mathbf}  
\begin{document}

\newcommand\mycommfont[1]{\footnotesize\rmfamily\textcolor{blue}{#1}}
\usetikzlibrary{arrows.meta}
\usetikzlibrary{positioning}
\tikzstyle{decision} = [diamond, draw, fill=blue!20, 
    text width=6em, text badly centered, node distance=3cm, inner sep=0pt]
\tikzstyle{block} = [rectangle, draw, fill=gray!10, 
    text width=10em, very thick, text centered, rounded corners, minimum height=2.2em]
\tikzstyle{line} = [draw, -{latex[scale=15.0]}]
\tikzstyle{cloud} = [draw, ellipse,fill=red!20, node distance=3cm,
    minimum height=2em]
\setlength{\fboxrule}{1pt}
\setlength{\fboxsep}{0pt}

\newcommand{\cmark}{\ding{51}}%
\newcommand{\xmark}{\ding{55}}%

\newtheorem{innercustomthm}{Theorem}
\newenvironment{customthm}[1]
  {\renewcommand\theinnercustomthm{#1}\innercustomthm}
  {\endinnercustomthm}

\newtheorem{innercustomprop}{Proposition}
\newenvironment{customprop}[1]
  {\renewcommand\theinnercustomprop{#1}\innercustomprop}
  {\endinnercustomprop}

\newtheorem{definition}{Definition}[section]
\newtheorem{prop}{Proposition}[section]
\newtheorem{theorem}{Theorem}[section]

\title{\LARGE \bf Unifying Scene Representation and Hand-Eye Calibration\\ with 3D Foundation Models 
}
\author{Weiming Zhi \and Haozhan Tang \and Tianyi Zhang \and Matthew Johnson-Roberson
\thanks{The authors are with the Robotics Institute, Carnegie Mellon University. Correspondence to \url{wzhi@andrew.cmu.edu}.}}
\maketitle

\begin{abstract}
Representing the environment is a central challenge in robotics, and is essential for effective decision-making. Traditionally, before capturing images with a manipulator-mounted camera, users need to calibrate the camera using a specific external marker, such as a checkerboard or AprilTag. However, recent advances in computer vision have led to the development of \emph{3D foundation models}. These are large, pre-trained neural networks that can establish fast and accurate multi-view correspondences with very few images, even in the absence of rich visual features. This paper advocates for the integration of 3D foundation models into scene representation approaches for robotic systems equipped with manipulator-mounted RGB cameras. Specifically, we propose the Joint Calibration and Representation (JCR) method. JCR uses RGB images, captured by a manipulator-mounted camera, to simultaneously construct an environmental representation and calibrate the camera relative to the robot's end-effector, in the absence of specific calibration markers. The resulting 3D environment representation is aligned with the robot's coordinate frame and maintains physically accurate scales. We demonstrate that JCR can build effective scene representations using a low-cost RGB camera attached to a manipulator, without prior calibration.
\end{abstract}

% \begin{IEEEkeywords}
% Motion Generation, Dynamical systems
% \end{IEEEkeywords}

\section{Introduction}
The manipulator-mounted camera setup, where the camera is rigidly attached to the manipulator, enables the robot to actively perceive its environment and is a common setup for robot manipulation. The robot needs to extract a concise representation of the physical properties of the environment from the collected data, enabling it to operate safely and make informed decisions. Compared to fixed cameras, manipulator-mounted cameras allow the robot system to adjust its viewing pose to reduce occlusion and obtain measurements at diverse angles and distances. However, manipulator-mounted cameras also come with their challenges --- the camera must be calibrated before collecting data from the environment. Specifically, to obtain scene representations that the robot can plan within, it is important to transform the representation into the reference frame of the robot base. This process of finding the camera pose relative to the end of the manipulator, or \emph{end-effector}, is known as \emph{hand-eye calibration} \cite{Tsai1988ANT}. 

Classical hand-eye calibration is an elaborate procedure that requires the camera to move to a diverse dataset of poses and record multiple images of an external calibration marker, usually a checkerboard or an AprilTag \cite{AprilTag}. Then, the rigid body transformation between the camera and the end-effector can be computed. This complex procedure can be a hurdle for non-experts since it necessitates the creation of dedicated markers and the collection of a new dataset each time the camera is recalibrated.

Deep learning approaches have driven recent advances within the computer vision community. This has led to the emergence of large pre-trained models, such as DUSt3R \cite{Wang2023DUSt3RG3}, which greatly outperform classical approaches for multi-view problems. These models are trained on large datasets and intended as \emph{plug-and-play} modules to facilitate a wide range of downstream tasks. We describe these models as \emph{3D Foundation Models} \cite{Bommasani2021FoundationModels} and advocate for their integration in robot camera calibration and scene representation.   

% \begin{figure}[t]
% \centering
%     \includegraphics[width=\linewidth]{img_fig1.pdf}
%     \caption{Our presented JCR method jointly calibrates the camera and builds environment representations (which can capture occupancy, color, and segmentation classes), from RGB images captured on a manipulator-mounted camera. Crucially, no external markers, such as AprilTags, are required for the calibration. The representations are of true scale and in the coordinate frame of the robot.}\label{fig:overview}
% \end{figure}

\begin{figure}[t]
\centering
    \includegraphics[width=\linewidth]{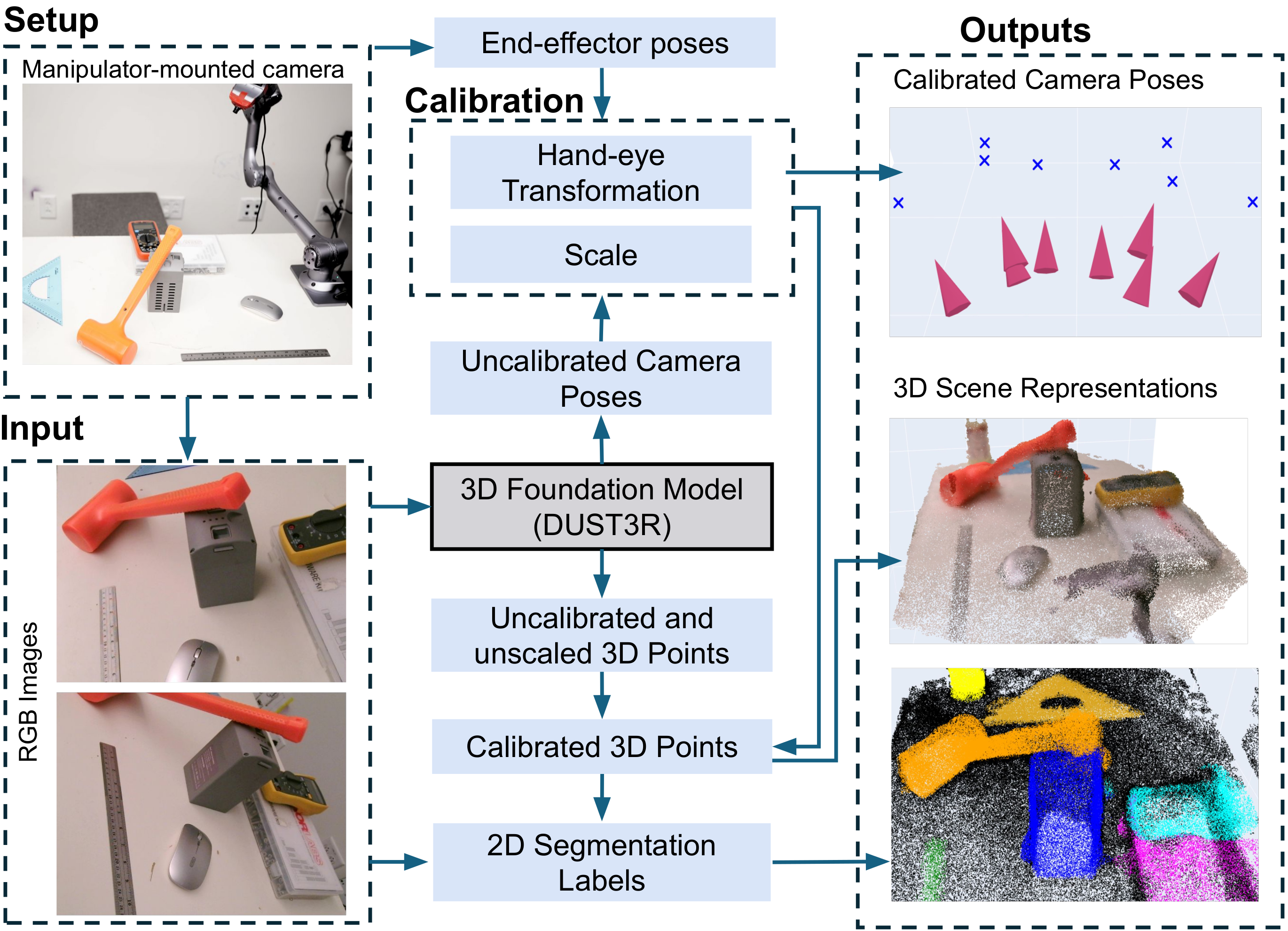}
    \caption{Our presented JCR method jointly calibrates the camera and builds environment representations (which can capture occupancy, colour, and segmentation classes), from RGB images captured on a manipulator-mounted camera. Crucially, no external markers, such as AprilTags \cite{AprilTag}, are required for the calibration. The representations are of true scale and in the coordinate frame of the robot.}\label{fig:overview}
\end{figure}

In this paper, we contribute the \emph{Joint Calibration and Representation (JCR)} method. JCR leverages a 3D foundation model to jointly conduct hand-eye calibration and construct a scene representation in the coordinate frame of the manipulator's base from a small set of images, collected from an RGB camera mounted on the manipulator. Previous approaches which use manipulator-mounted sensors require capturing images of external markers, which are then used to perform calibration. To the best of our knowledge, \textbf{the proposed approach is the first to simultaneously calibrate the camera and build a scene representation from the same set of images captured by a manipulator-mounted camera.} We obtain a model of our environment in the robot's coordinate frame, without any \emph{a priori} calibration, external markers, or depth readings. The constructed scene representation is a continuous model which can be used for collision-checking in subsequent motion planning. We validate the robustness of our approach using a variety of collected real-world datasets, collected from a low-cost camera mounted on a 6-DOF manipulator. A diagrammatic overview of JCR is presented in \Cref{fig:overview}. 

The remainder of the paper is organized as follows: We begin by discussing related work in \Cref{sec:related} and then introduce the necessary background on 3D foundation models in \Cref{sec:preliminaries}. We detail the technical aspects of the Joint Calibration and Representation (JCR) method in \Cref{sec:JCR}, follow with empirical evaluations of JCR in \Cref{sec:experiments}, and conclude by summarizing our findings and outlining future research directions in \Cref{sec:conclusions}.

\section{Related Work}\label{sec:related}
\textbf{Scene Representation:} 
Early work on representing environments in robotics typically recorded environment properties in discretized cells, with the most notable approach being Occupancy Grid Maps \cite{OccupancyGridMaps}. Distance-based representations have also been applied for robotics tasks \cite{SDF, kinect_fusion} to check for collisions. Advances in machine learning have motivated the development of continuous representation methods, which functionally represent structure in the environment. For example, by using Gaussian processes \cite{SimonGPOM}, kernel regression \cite{HilbertMaps, sptemp}, Bayesian methods \cite{HM,Senanayke:2017, wright2024vprism}, and neural networks \cite{Park_2019_CVPR}. Deep learning approaches to directly operate on point clouds \cite{pointNet, pointnet_plusplus} have also been developed. Methods to ingest point clouds directly for robot planning have also been explored \cite{GeoFab_gloabL_opt}. Concurrently, in the computer vision community, there has been an effort to create photo-realistic scene representations. These approaches include Neural Radiance Fields (NeRFs) \cite{mildenhall2020nerf} and subsequent variants \cite{mueller2022instant, zhang2024darkgs}. They rely on obtaining an initial solution from Structure-from-Motion methods \cite{schoenberger2016sfm}, and train an implicit model that matches the environment's appearance. 

\textbf{Hand-eye Calibration:} Hand-eye calibration is a well-studied problem with geometric solutions \cite{Tsai1988ANT, hand-eye, Park+Martin-1994} developed to solve for the transformation when some external calibration marker, such as a checkerboard, is provided. A recent learning approach for hand-eye calibration is presented in \cite{pmlr-v164-valassakis22a}, but requires a part of the robot's gripper to be visible in the camera view. Additionally, there exist methods in end-to-end policy learning \cite{decision_trans, implicitBC} which directly train for actions from the camera images, and do not require hand-eye calibration. However, unlike our method, these methods are unable to construct an environment model which can then be used for collision-checking in downstream motion planning and decision-making \cite{lavalle1998_RapiRand, PDMP, diagrammaticlearning}.

\textbf{Pre-trained Models:}
The machine learning community has made considerable efforts to develop large-scale models trained on extensive web data, resulting in significant advancements in large deep learning models for natural language processing \cite{touvron2023llama} and computer vision \cite{clip}. In particular, \cite{Wang2023DUSt3RG3, depthanything} are pre-trained models that can be applied to 3D tasks. These large pre-trained models are known as \emph{foundation models} \cite{Bommasani2021FoundationModels} and are typically treated as back-boxes whose output is used in subsequent downstream tasks. Although the outputs of these models generally require further processing before applying them to robotics tasks, there has been widespread interest in incorporating foundation models within robot systems \cite{Firoozi2023FoundationMI}.

\section{Prelminaries: 3D Foundation Models for Dense Reconstruction}\label{sec:preliminaries}
Traditional methods for 3D tasks such as Structure-from-Motion \cite{schoenberger2016sfm} or multi-view stereo \cite{stereo_tut} depend on identifying visual features over a set of images to construct corresponding 3D structures. On the other hand, pre-trained models, such as \emph{Dense Unconstrained Stereo 3D Reconstruction} (DUSt3R), have been trained on large datasets and can identify correspondences over a set of images without strong visual features. Throughout this work, we use DUSt3R \cite{Wang2023DUSt3RG3} as the foundation model and follow its conventions. Here we shall briefly outline how the foundation model estimates relative camera poses from RGB images and more details can be found in \cite{Wang2023DUSt3RG3}. 

\textbf{Pairwise Pixel Correspondence:} Suppose we have a pair of RGB images with width $W$ and height $H$, i.e. $I_{1},I_{2} \in \mathbb{R}^{W\times H\times 3}$, our foundation model can produces \emph{pointmaps}, $X^{1,1}, X^{1,2}\in\mathbb{R}^{W\times H\times 3}$. Pointmaps assign each pixel in the 2D image to its predicted 3D coordinates and are \textbf{critically in the same coordinate frame of $I_{1}$}. Confidence maps, $C^{1,1}, C^{1,2}\in\mathbb{R}^{W \times H}$, for each of the pointmaps are also produced. These indicate the uncertainty of the foundation model's prediction for each point. By finding the nearest predicted 3D coordinates of each pixel in the pointmap with the coordinates of other pointmap, we can find dense correspondences between pixels in the image pair, without handcrafted features. 

\textbf{Recovering Relative Camera Poses:} We optimize to globally align the pairwise pointmaps predicted by the foundation model to recover the relative camera poses corresponding to \emph{a set of images}. For a set of $N$ images, we have the cameras $n = 1,\dots,N$ and possible image pairs with indices $(n,m)\in \varepsilon$, where $m=1,\ldots,N$ and $m\neq n$. For each pair, the foundation model gives us:
(1) Pointmaps in $X^{n,n}, X^{n,m}\in \mathbb{R}^{W \times H \times 3}$ in the frame of $I_{n}$; 
(2) Corresponding confidence maps $C^{n,n}, C^{n,m}\in \mathbb{R}^{W \times H}$.
With these, we seek to optimize to find:
(1) For each of the $N$ images, a pointmap in global coordinates $\bar{X}^{n}$;
(2) A rigid transformation described $P_{n}\in\mathbb{R}^{3 \times 4}$ and factor $\sigma_{n}>0$.

Intuitively, the same transformation should be able to align both images in each pair to their equivalents in the global coordinate. We can then minimize the distance between the transformed pointmaps and the predicted pointmaps in global coordinates: 
\begin{equation}
\min_{\hat{X},P,\sigma}\sum_{(n,m)}
\sum_{i\in (n,m)}\sum_{(w,h)} C^{n,i}_{w,h}\lvert\lvert \hat{X}_i-\sigma_{e}P_{e}X^{n,e}_{w,h}\lvert\lvert_{2}.\label{eqn:opt}
\end{equation}
Here, $(n,m)\in \varepsilon$ denotes the pairs, $i$ iterates through the two images in each pair, and $(w,h)$ iterates through each pixel in the image. \Cref{eqn:opt} can be optimized efficiently via gradient descent. With the pointmaps over a set of images in the same coordinate frame, we can extract the set of camera poses $P$ and the globally aligned pointmaps $\hat{X}$ over the set of images, and form a pointset representation of the environment. %which form a dense reconstruction of the scene.

However, the obtained outputs of the foundation model cannot be directly used to build representations for robots to operate in. We need the outputs of the foundation model to be in the robot's coordinate frame. Additionally, the 3D foundation models typically cannot recover physically accurate scale. Here, $\sigma_{n}$ does not correspond to physical scales, and we need to re-scale distances by the unknown model-to-reality factor to be physically accurate.

\begin{algorithm}[t]
%\scriptsize
% \normalsize
%\small
\caption{{Joint Calibration and Representation}}\label{alg:jcr}
\SetKwInOut{Input}{input}
\SetKwInOut{Output}{output}
\SetKwInOut{Initalise}{initalise}
\SetAlgoLined
\newcommand{\lIfElse}[3]{\lIf{#1}{#2 \textbf{else}~#3}}
\SetKwFor{DoParallel}{Do in Parallel for each (}{) $\lbrace$}{$\rbrace$}
\Input{End-effector poses $\{E_{i},\ldots,E_{N}\}$, Captured images $\{I_{1},\ldots,I_{N}\}$, 3D foundation model ($\mathtt{3DFM}$), Neural network $f_{\theta}$}
$\{P_{i}\}_{i=1}^{N}, \{\hat{X}_{i}\}_{i=1}^{N}, \{C_{i}\}_{i=1}^{N}\leftarrow \mathtt{3DFM}(\{I_{1},\ldots,I_{N}\})$; {\color{blue} Input images into foundation model to find uncalibrated camera poses, pointmaps, confidence.}\\
$\bigg\{
\begin{array}{l}
T_{E_{i}}^{E_{i+1}} = E_{i+1}(E_{i})^{-1}, \\
T_{P_{i}}^{P_{i+1}} = P_{i+1}(P_{i})^{-1}.
\end{array}
\text{for } i\!=\!1,\ldots,N-1$; {\color{blue}Rearranging \Cref{eq:get_T}.}\\
Obtain ${R_{c}^{e}}^{*}$ via \Cref{eq:solve_R};\\
Solve for ${\bb{t}_{c}^{e}}^{*},\lambda$ via \Cref{eq:srp};\\
Obtain $\{\bb{x}_{i}\}_{i=1}^{N_{pc}}$ by filtering confident 3D points from $\{\hat{X}_{i}\}_{i=1}^{N}$ with $\{C_{i}\}_{i=1}^{N}$;\\
$\bb{\bar{x}}_{i}=E^{-1}{T_{c}^{e}}^{*}(\lambda^{*}\bb{x}_{i})$, where ${T_{c}^{e}}^{*}=\begin{bmatrix}
        {R_{c}^{e}}^{*} & {\bb{t}_{c}^{e}}^{*}\\
        0\quad 0\quad 0 & 1
    \end{bmatrix}$; {\color{blue} Transform $\bb{x}$ into robot's frame via \Cref{eq:transform_frame}.} \\
Train $f_{\theta}$ via \Cref{eq:train_occ}.\\
\Output{Occupancy Representation $f_{\theta}$, and transformation ${T_{c}^{e}}^{*}$.}
\end{algorithm}

\section{Joint Calibration and Representation}\label{sec:JCR}
% In the follow section, we shall briefly outline the problem setup. Then, we detail how the outputs of the foundation model can be used to jointly calibrate the camera and construct a concise environment representation that is in the robot's frame.
\subsection{Overview}
\begin{figure}[t]
\centering
  \begin{subfigure}{.95\linewidth}
    \centering
    \fbox{\includegraphics[width=.47\linewidth]{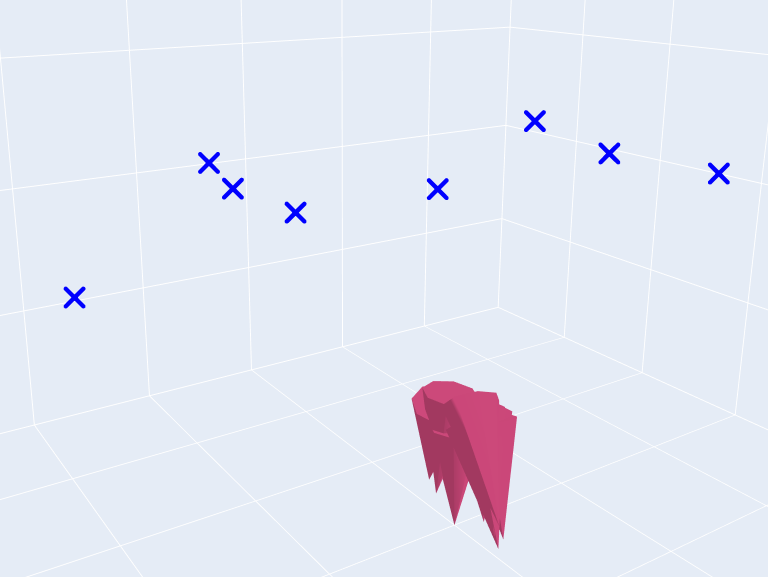}}%
    \hspace{0.05em}
    \fbox{\includegraphics[width=.47\linewidth]{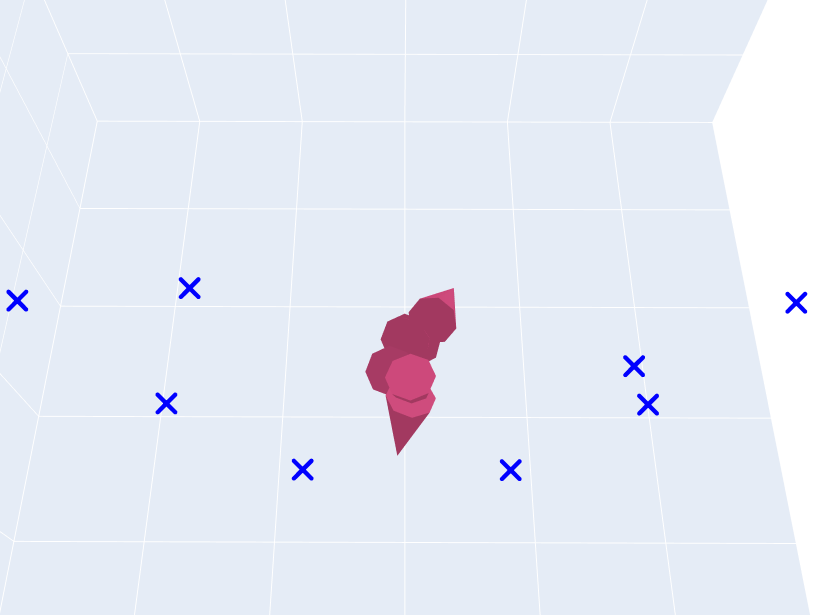}}%
    \caption{Before Calibration}
  \end{subfigure}%
\vspace{0.5em}
  \begin{subfigure}{.95\linewidth}
    \centering
    \fbox{\includegraphics[width=.47\linewidth]{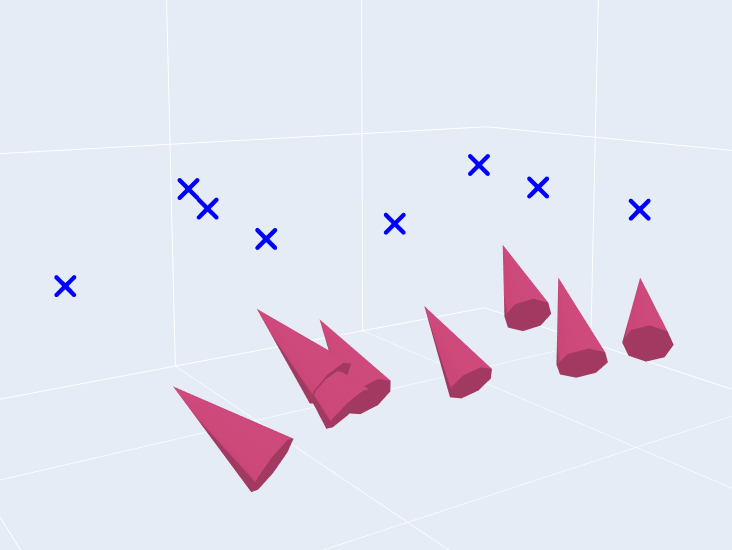}}%
    \hspace{0.05em}
    \fbox{\includegraphics[width=.47\linewidth]{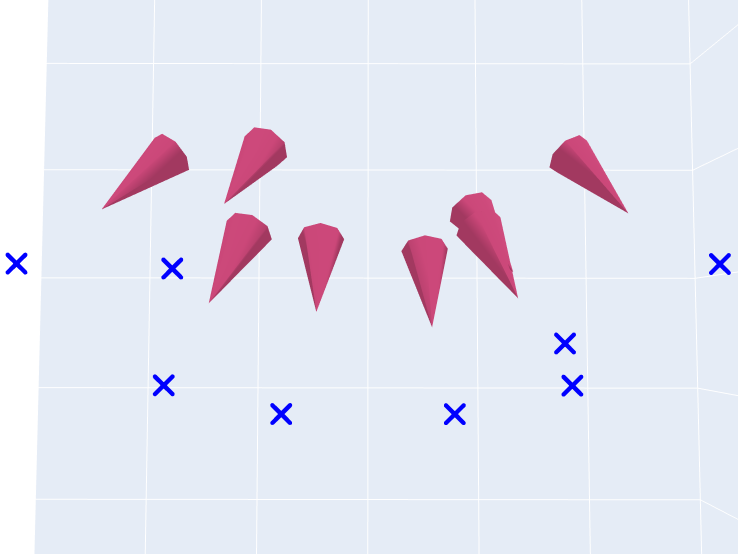}}%
    \caption{After Calibration and Rescaling}
  \end{subfigure}

\caption{An example of end-effector positions (shown as blue crosses) and corresponding camera poses (given by cones, with cone bases indicating camera orientation), before and after calibration and rescaling. The camera poses from the foundation model (subfigure a) do not correspond to the physical scale nor align with the end-effector. This is resolved by solving \Cref{eq:srp} (subfigure b).}\label{fig:Calibration}
\end{figure}

We tackle the problem setup of a manipulator with an inexpensive RGB camera rigidly mounted on the manipulator. Here, we do not require the mounted camera to be \emph{calibrated}, that is, the camera pose relative to the end-effector is unknown. We control the end-effector manipulator to go to a small set of $N$ poses, $\{E_{1},\ldots, E_{N}\}$, and capture an image at each pose. This gives us a set of images $N$ of the environment, $\{I_{1},\ldots, I_{N}\}$, which can be inputted into the foundation model to obtain a set of aligned relative camera poses $\{P_{1},\ldots, P_{N}\}$ and pointmaps $\{\hat{X}_{1},\ldots, \hat{X}_{N}\}$ with respect to an arbitrary coordinate system and scale. \Cref{fig:Calibration} shows an example of before and after calibration and scaling. We observe that the relative camera poses from the foundation model are inconsistent with the end-effector poses and physical scaling. 

Here, JCR seeks to recover:
\begin{itemize}
\item The rigid transformation, $T_{c}^{e}$, from the frame of the mounted camera to that of the end-effector. 
\item A scale factor $\lambda$, that scales the foundation model's unscaled outputs to true physical scale.  
\item A representation of the environment in the robot's frame that identifies occupied space, and other properties of interest, such as segmentation classes and RGB colour.
\end{itemize}
Obtaining $T_{c}^{e}$ will enable us to efficiently incorporate new images from the camera into the robot's frame, while the environment representation is critical downstream to planning tasks. An algorithmic outline of JCR is presented in \Cref{alg:jcr}.

\subsection{Calibration With Foundation Model Outputs}
Here, we seek to solve for $T_{c}^{e}$ with the end-effector poses $\{E_{1},\ldots,E_{N}\}$ the predicted unscaled relative camera poses $\{P_{1},\ldots,P_{N}\}$. We shall consider transformations between subsequent end-effector poses $T_{E_{i}}^{E_{i+1}}$ and transformations between camera poses $T_{P_{i}}^{P_{i+1}}$, where
\begin{align}
\Bigg\{\begin{array}{l}
E_{i+1} = T_{E_{i}}^{E_{i+1}} E_{i} \\
P_{i+1} = T_{P_{i}}^{P_{i+1}} P_{i}
\end{array} \quad \text{for } i=1,\ldots,N-1.\label{eq:get_T}
\end{align}
As the foundation model does not recover absolute scale, we introduce a \emph{scale factor} $\lambda$. The transformation between scaled estimated camera poses as:
\begin{equation}
    T_{P_{i}}^{P_{i+1}}(\lambda)=\begin{bmatrix}
        R_{P_{i}}^{P_{i+1}} & \lambda \bb{t}_{P_{i}}^{P_{i+1}}\\
        0\quad 0\quad 0 & 1
    \end{bmatrix}\in\mathbf{SE}(3),
\end{equation}
where $R_{P_{i}}^{P_{i+1}}\in \mathbf{SO}(3)$ denotes the rotation component of $T_{P_{i}}^{P_{i+1}}$ and $\bb{t}_{P_{i}}^{P_{i+1}}\in\mathbb{R}^{3}$ denotes the translation. Here, we note that scaling the distances between predicted camera poses does not affect the rotation but scales the translation. 

The relationship between $T_{E_{i}}^{E_{i+1}}$, $T_{P_{i}}^{P_{i+1}}(\lambda)$ and the desired $T_{c}^{e}$ follows the matrix equation from classical hand-eye calibration \cite{Tsai1988ANT}:
\begin{equation}
T_{E_{i}}^{E_{i+1}}T_{c}^{e}=T_{c}^{e}T_{P_{i}}^{P_{i+1}}(\lambda), \label{eqn:cali}
\end{equation}
and we shall solve for the best fit $T_{c}^{e}$ and $\lambda$. We begin by solving for the rotational term $R_{c}^{e}$ by following \cite{Park+Martin-1994}, and considering the log map of $\mathbf{SO}(3)$ to its lie algebra ($\mathfrak{so}(3)$) where for some $R\in\mathbf{SO}(3)$,
\begin{align}
\omega=&\arccos(\frac{\mathrm{Tr}(R)-1}{2}), \\
\mathrm{LogMap}(R):=&\frac{\omega}{2\sin(\omega)}\begin{bmatrix}
R_{3,2}-R_{2,3}\\ R_{1,3}-R_{3,1} \\ R_{2,1}-R_{1,2}
\end{bmatrix} \in \mathfrak{so}(3).
\end{align}
Here, the subscripts indicate the elements in $R$, and $\mathrm{Tr}(\cdot)$ indicates the trace operator. Then, the best fit rotation ${R_{c}^{e}}^{*}$ can be found via:
\begin{align}
{R_{c}^{e}}^{*}&=(M^{\top}M)^{-\frac{1}{2}}M^{\top}, \label{eq:solve_R}\\
\text{where } M&=\sum_{i=1}^{N-1} \mathrm{LogMap}(R_{E_{i}}^{E_{i+1}})\otimes \mathrm{LogMap}(R_{P_{i}}^{P_{i+1}}),\nonumber
\end{align}
where $\otimes$ denotes the outer product and the matrix inverse square root can be computed efficiently via singular value decomposition. 

Next, we formulate a residual optimization problem to find the best-fit translation ${\bb{t}_{c}^{e}}^{*}$ and scale ${\lambda}^{*}$ by minimizing the residuals. We formulate the \emph{Scale Recovery Problem} (SRP):
\begin{align}
\text{SRP:} \quad \arg\min_{\bb{t}_{c}^{e},\lambda}&\sum^{N-1}_{i=1}\lvert\lvert C_{i}\bb{t}_{c}^{e} - \bb{d}_{i}(\lambda)\lvert\lvert_{2}^{2}, \label{eq:srp} \\
\text{where }  C_{i} = I-R_{E_{i}}^{E_{i+1}}&, \quad \bb{d}_{i}(\lambda)=\bb{t}_{E_{i}}^{E_{i+1}}-{R_{c}^{e}}^{*}(\lambda\bb{t}_{c}^{e}).
\end{align}
For a fixed $\lambda$, the translation solution admits the closed-form solution 
\begin{align}
{\mathbf{t}_c^e}^{*} (\lambda) = (C^T C)^{-1} C^T \mathbf{d}(\lambda). 
\end{align}
Here, $C$ is a matrix composed of $C_{i}$ stacked vertically and $\mathbf{d}(\lambda)$ is a vector composed of the corresponding $\mathbf{d}_{i}(\lambda)$ values. We can then optimize $\lambda$ numerically to obtain can obtain the optimal scale factor $\lambda^{*}$. The entire camera to end-effector transformation can then be obtained via
\begin{equation}
    {T_{c}^{e}}^{*}=\begin{bmatrix}
        {R_{c}^{e}}^{*} & {\bb{t}_{c}^{e}}^{*}\\
        0\quad 0\quad 0 & 1
    \end{bmatrix}\in\mathbf{SE}(3).
\end{equation}

\begin{table*}[t]
\begin{adjustbox}{width=0.99\textwidth,center}
\begin{tabular}{@{}ll|rrr|rrr|rrr@{}}
\toprule[1pt]\midrule[0.3pt]

\multicolumn{1}{l}{} & \multicolumn{1}{l}{} & \multicolumn{1}{l}{} & \multicolumn{1}{l}{Light Tabletop} & \multicolumn{1}{l}{} & \multicolumn{1}{l}{} & \multicolumn{1}{l}{Light Tabletop} & \multicolumn{1}{l}{} & \multicolumn{1}{l}{} & \multicolumn{1}{l}{Dark Tabletop} & \multicolumn{1}{l}{} \\ 
\multicolumn{1}{l}{} & \multicolumn{1}{l}{} & \multicolumn{1}{l}{} & \multicolumn{1}{l}{(8 items)} & \multicolumn{1}{l}{} & \multicolumn{1}{l}{} & \multicolumn{1}{l}{(7 items)} & \multicolumn{1}{l}{} & \multicolumn{1}{l}{} & \multicolumn{1}{l}{} & \multicolumn{1}{l}{} \\ 
\midrule
      &    Images Provided                   & 10 images            & 12 images                            & 15 images            & 10 images            & 12 images                            & 15 images            & 10 images            & 12 images                 & 15 images          \\ \midrule 
                     & Converged            & $\bm{\checkmark}$                    & $\bm{\checkmark}$                                    & $\bm{\checkmark}$                    & $\bm{\checkmark}$                    & $\bm{\checkmark}$                                    & $\bm{\checkmark}$                    & $\bm{\checkmark}$                    & $\bm{\checkmark}$                         & $\bm{\checkmark}$                    \\
Ours                 & Residual $\delta_\bb{t}$             & 0.0420                               & 0.0419                                               & 0.0396                               & 0.0208                               & 0.0317                                               & 0.0357                               & 0.0310                               & 0.0536                                    & 0.0414                              \\
                     & Residual $\delta_R$               & 0.0655                               & 0.0657                                               & 0.0513                               & 0.0519                               & 0.0623                                               & 0.0701                               & 0.0732                               & 0.0742                                    & 0.0818                              \\
                     & No. of Poses             & 10                                   & 12                                                   & 15                                   & 10                                   & 12                                                   & 15                                   & 10                                   & 12                                        & 15                                   \\\midrule
                     & Converged            & \textbf{\texttimes}                 & \textbf{\texttimes}                                 & \textbf{\texttimes}                 & $\bm{\checkmark}$                    & $\bm{\checkmark}$                                    & $\bm{\checkmark}$                    & \textbf{\texttimes}                 & \textbf{\texttimes}                      & $\bm{\checkmark}$                    \\
COLMAP \cite{schoenberger2016sfm}               & Residual $\delta_\bb{t}$              & NA                                   & NA                                                   & NA                                   & 0.0412                               & 0.0412                                               & 0.0469                               & NA                                   & NA                                        & 0.0454                              \\
      + Calibration           & Residual $\delta_R$                   & NA                                   & NA                                                   & NA                                   & 1.27                                 & 1.27                                                 & 0.0662                               & NA                                   & NA                                        & 0.0503                              \\
                     & No. of Poses             & 2                                    & 2                                                    & 2                                    & 5                                    & 5                                                    & 10                                   & 4                                    & 4                                         & 10                                   \\ \midrule[0.3pt]\bottomrule[1pt]
\end{tabular}

\end{adjustbox}
\caption{We evaluate our JCR against estimating camera poses with COLMAP and then run calibration. We observe that, especially when the number of images is low, COLMAP can only estimate a few camera poses, resulting in divergence and large residuals. Our method can accurately run hand-eye calibration even when a low number of images are provided.}\label{table: res_cali}
\end{table*}

\subsection{Map Construction with Foundation Model Outputs}
Next, we seek to build representations of the environment with the output of the foundation model: (1) a set of aligned pointmaps $\{\hat{X}_{1},\ldots, \hat{X}_{N}\}$ with associated confidence maps $\{C_{1},\ldots,C_{N}\}$. From there, we can set a confidence threshold and filter out points in each $\hat{X}$ to be below the threshold, and obtain a 3D point cloud $\{\bb{x}_{i}\}_{i=1}^{N_{pc}}$, which is in the coordinate frame of some camera pose $P$, with the end-effector pose $E$. We transform the point cloud from the coordinate frame of the camera to that of the robot and adjust the scale to match the real world via:
\begin{align}
\bb{\bar{x}}_{i}=E^{-1}{T_{c}^{e}}^{*}(\lambda^{*}\bb{x}_{i}), && \text{for }i=1,\ldots,N_{pc}, \label{eq:transform_frame}
\end{align}
where $\bb{\bar{x}}_{i}\in\mathbb{R}^{3}$ are now in the robot's frame, and ${T_{c}^{e}}^{*}$, $\lambda^{*}$ are solutions of \Cref{eq:srp}. 

\textbf{Representing Occupancy:} The \emph{occupancy} information, i.e. whether a coordinate is occupied or not, is useful for planning tasks in the environment. Here, we use a small neural network $f_\theta$ to learn a continuous and \emph{implicit} model of occupancy. It assigns each spatial coordinate a probability of being occupied. We take a Noise Contrastive Estimate (NCE) \cite{NCE} approach and minimize the binary cross-entropy loss (BCELoss) \cite{Bishop:2006}, with $\bb{\bar{x}}_{i}$ as positive examples and uniformly drawing negative examples $\bb{\bar{x}}^{neg}_{i}$. Similar to NeRF models \cite{mildenhall2020nerf}, we apply sinusoidal position embedding $\phi$ on the positions before inputting the encoding to the network. Our loss function is given by:      
\begin{align}
L(\theta)=&BCELoss(\{f_{\theta}(\phi(\bb{\bar{x}}_{i}))\}_{i=1}^{N_{pc}},\{f_{\theta}(\phi(\bb{\bar{x}}_{i}^{neg}))\}_{i=1}^{N_{pc}}),\nonumber\\ 
&\text{where}\quad\bb{\bar{x}}_{i}^{neg}\sim U(\bb{\bar{x}}_{min}^{neg},\bb{\bar{x}}_{max}^{neg}), \label{eq:train_occ}
\end{align}
where $U(\bb{\bar{x}}_{min}^{neg},\bb{\bar{x}}_{max}^{neg})$ denotes a uniform distribution between boundaries $\bb{\bar{x}}_{min}^{neg}$, $\bb{\bar{x}}_{max}^{neg}$. We can then train the fully connected neural network, with a Sigmoid output layer, by optimizing $f_{\theta}$ with respect to parameters $\theta$. We can then query the trained neural network to predict whether a region of space is occupied. We can further build representations that capture properties in the occupied spatial coordinates. 

\textbf{Representing Segmentation:} After querying for occupancy in the environment, we may also wish to capture the segmentation of the 3D space into semantically meaningful parts. For example, we may want to be able to differentiate objects on the tabletop in the representation. As pointmaps outputted by the foundation model correspond to pixels in the RGB images, we can run 2D segmentation on the images (for example, using pre-trained models such as Segment Anything \cite{kirillov2023segany}). Provided segmentation labels over each pixel in the set of provided images, we can assign a segmentation class to each 3D point. We arrive at $\{\bb{\bar{x}}_{i}, {y}^{seg}_{i}\}_{i=1}^{N_{pc}}$, where ${y}^{seg}_{i}$ are segmentation class labels that correspond to each $\bb{\bar{x}}$. Then, we can treat the representation construction as a multi-class classification problem, and apply a positional encoding $\phi$ and optimize the multi-class cross-entropy \cite{Bishop:2006} loss:  
\begin{align}
L(\theta)=CrossEntropy(\{f_{\theta}(\phi(\bb{\bar{x}}_{i}))\}_{i=1}^{N_{pc}},\{{y}^{seg}_{i}\}_{i=1}^{N_{pc}}).
\end{align}
Here $f_{\theta}$ will be a fully connected neural network with a Softmax activation output layer.

\textbf{Representing Continuous Properties: } We can learn a neural network model $f_{\theta}$ to assign potentially multi-dimensional continuous properties to spatial coordinates by simply regressing onto labels. We can, for example, assign continuous colour values to points in the scene. As the pointmaps from the foundation model correspond pixel-wise to input images, we can obtain a 3-dimensional RGB colour label for each point, giving us a dataset $\{\bb{\bar{x}}_{i}, \bb{y}^{rgb}_{i}\}_{i=1}^{N_{pc}}$. Then, we can optimize the MSE loss:
\begin{align}
L(\theta)=MSELoss(\{f_{\theta}(\phi(\bb{\bar{x}}_{i}))\}_{i=1}^{N_{pc}},\{\bb{y}^{rgb}_{i}\}_{i=1}^{N_{pc}}).
\end{align}
After training, we can check for occupied regions from the occupancy representation, and then predict the colour assigned of the coordinates via a forward pass of $f_{\theta}$.

\begin{figure}[t]
\centering
\fbox{\includegraphics[angle=90,width=0.29\linewidth]{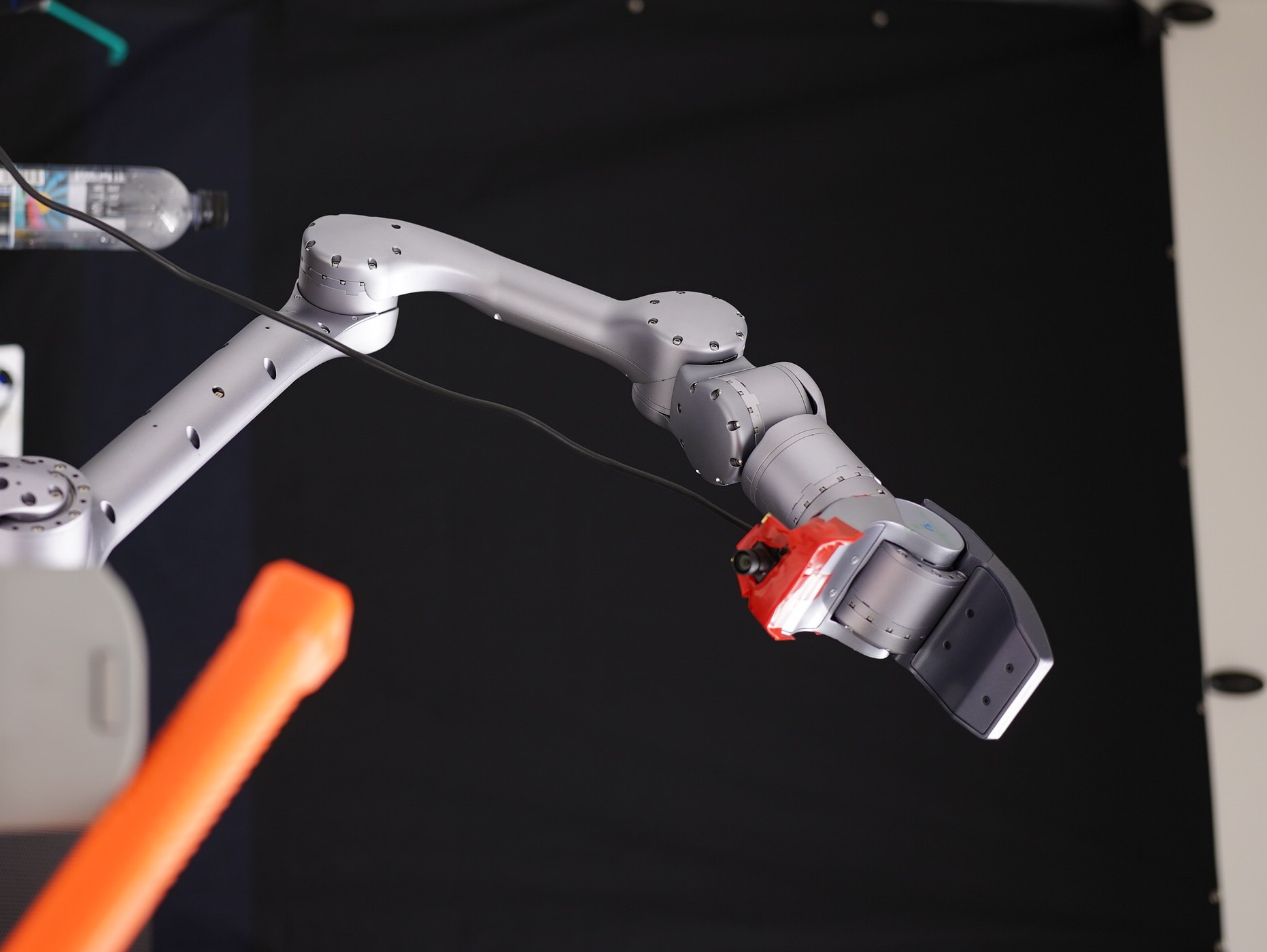}}%
\hfill
\fbox{\includegraphics[width=0.685\linewidth]{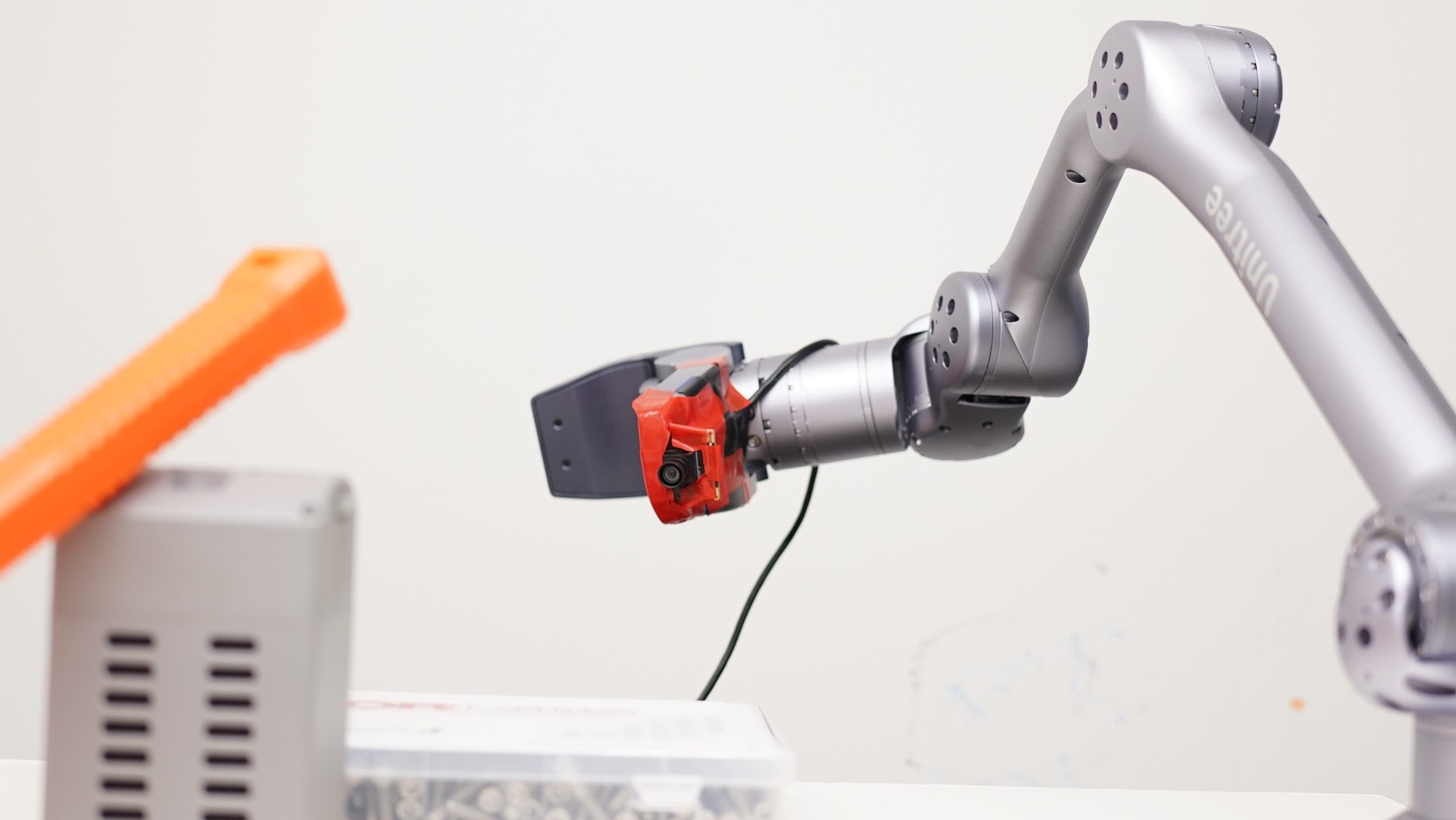}}
\caption{We attach an inexpensive and small RGB camera to our manipulator and take images at different end-effector poses. The transformation from the camera to the end-effector is not known in advance and is solved by our calibration.}\label{fig:setup}
\end{figure}

\section{Empirical Evaluations}\label{sec:experiments}
In this section, we evaluate the quality of the proposed Joint calibration and Representation (JCR) method to calibrate the camera with respect to the manipulator end-effector, as well as to build environment representations. We attach an inexpensive USB webcam (with an estimated retail cost of $10$ USD), which captures low-resolution RGB images, onto a Unitree Z1 6 degrees-of-freedom manipulator. We illustrate our robot setup in \Cref{fig:setup}. Compared to depth cameras, RGB cameras are smaller in size and lower in cost, making our vision-only setup an attractive option. The foundation model used within the joint calibration and representation framework is DUSt3R \cite{Wang2023DUSt3RG3}, using pre-trained weights for image inputs with width $512$ pixels. Here, the questions we seek to answer are:
\begin{enumerate}
    \item Can JCR, with foundation models, enable \emph{image efficient} hand-eye calibration, when the number of images provided is low?
    \item Can we recover the scale accurately by solving the scale recovery problem \ref{eq:srp}, such that our representation's sizes match the physical world?
    \item Can high-quality environment representations be built with JCR?    
\end{enumerate}

\begin{figure}[t]
    \centering
    \begin{subtable}[b]{0.68\linewidth}
        \centering
        \begin{adjustbox}{width=0.99\linewidth,center}
        \begin{tabular}{l|rrrr}
        \toprule[1pt]\midrule[0.3pt]
                  & \multicolumn{1}{l}{Tape} & \multicolumn{1}{l}{Box} & \multicolumn{1}{l}{Mug} & \multicolumn{1}{l}{Toolbox} \\
                  \midrule
        8 Images  & 7.8\%                    & 8.6\%                   & 1.2\%                   & 1.2\%                       \\
        10 Images & 2.5\%                    & 2.9\%                   & 3.1\%                   & 0.7\% \\
        \midrule[0.3pt]\bottomrule[1pt]
        \end{tabular}
        \label{subtable:scale}

        \end{adjustbox}
        \caption{The percentage error of object heights. We observe that with 10 images, the recovered heights of each object all have errors that are at most $3.1\%$, indicating accurate scale recovery.}\label{table:scale}
    \end{subtable}%
    \hfill
    \begin{subfigure}[b]{0.27\linewidth}
        \centering
        \fbox{\includegraphics[width=\linewidth]{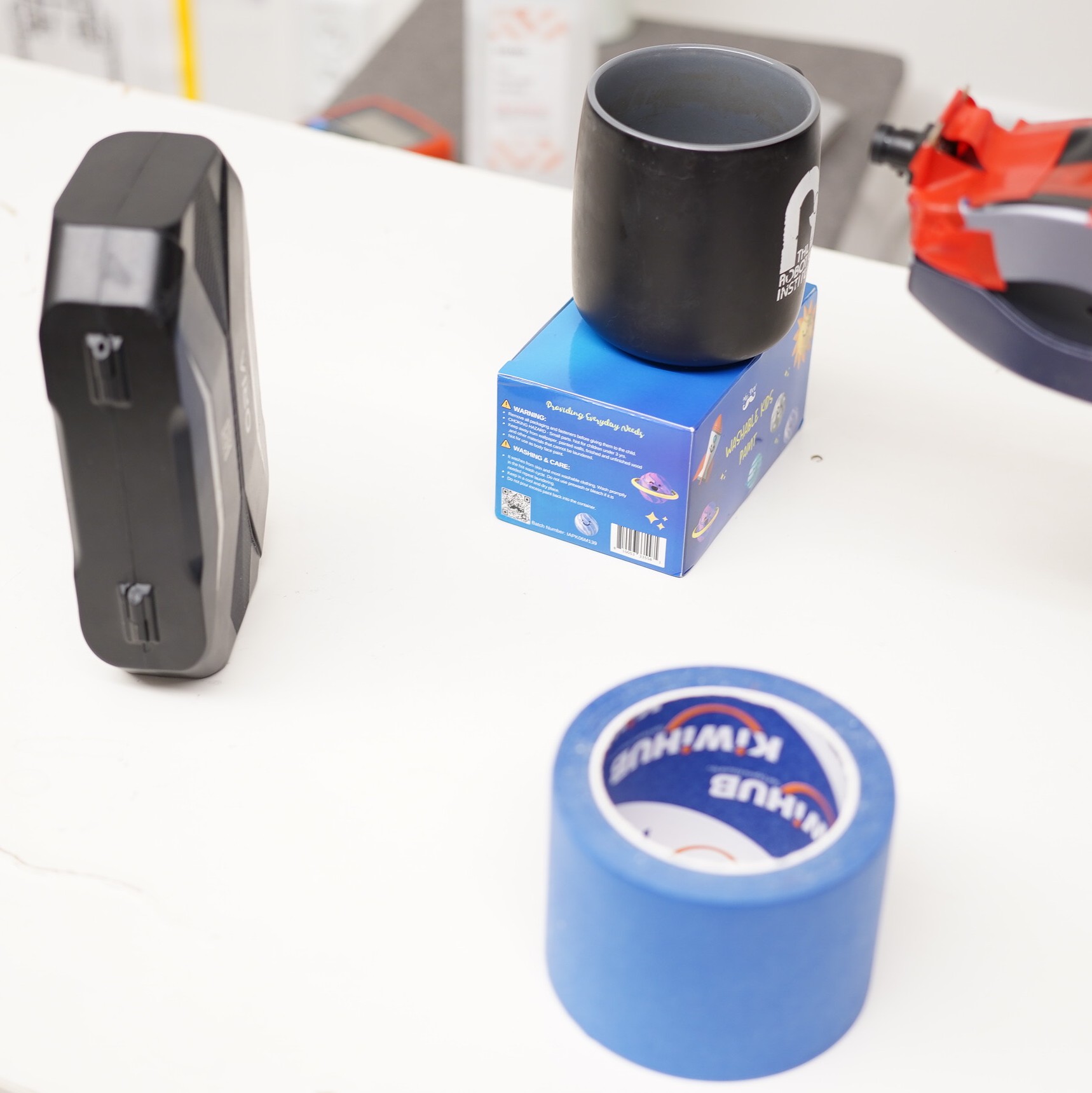}}
        \caption{Measured items.}
        \label{fig:items}
    \end{subfigure}

    \caption{Percentage error between true heights and recovered heights.}
\end{figure}

\begin{figure*}[t]
  \centering

  \begin{subfigure}{.325\textwidth}
    \centering
    \fbox{\includegraphics[width=.495\linewidth]{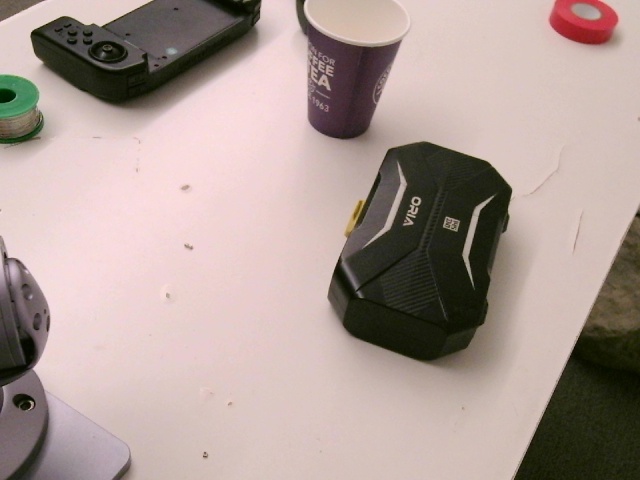}}%
    \fbox{\includegraphics[width=.495\linewidth]{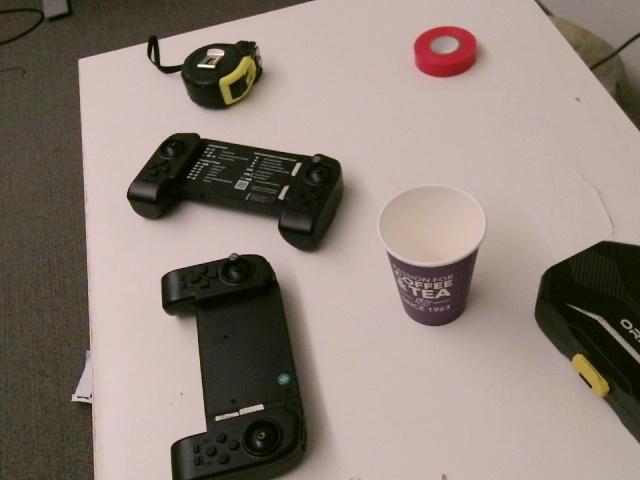}}\\
    \fbox{\includegraphics[width=.495\linewidth]{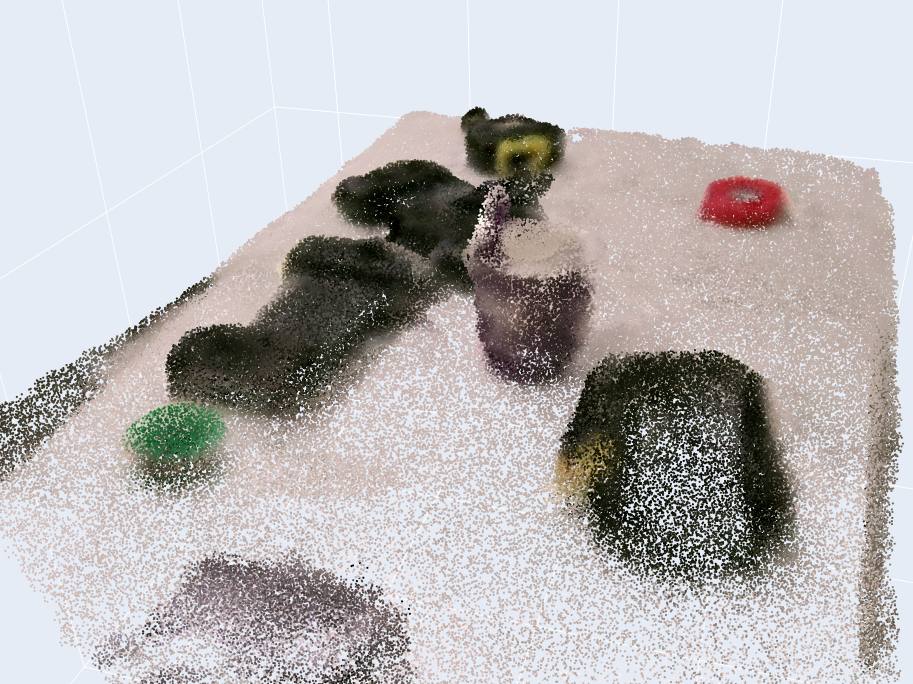}}%
    \fbox{\includegraphics[width=.495\linewidth]{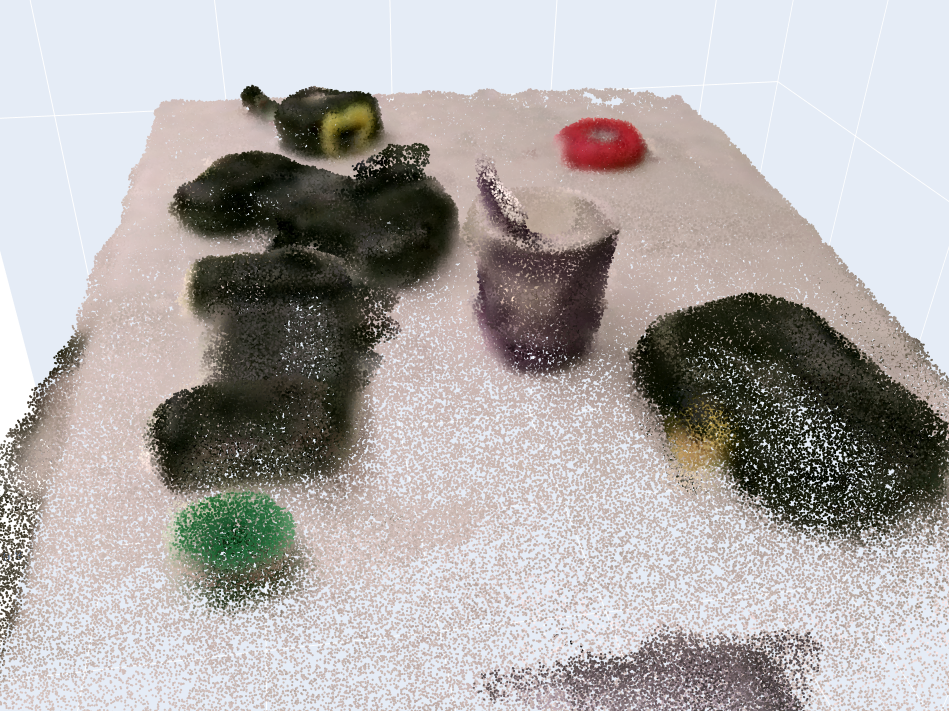}}%
    \caption{Light Tabletop 7 items}
  \end{subfigure}
  \hfill
\begin{subfigure}{.325\textwidth}
    \centering
    \fbox{\includegraphics[width=.495\linewidth]{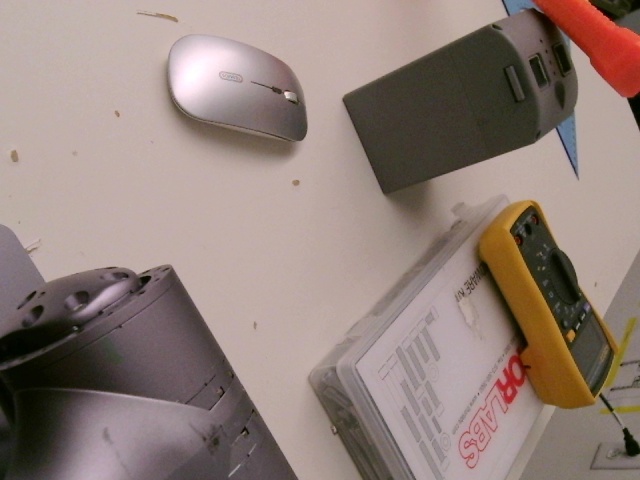}}%
    \fbox{\includegraphics[width=.495\linewidth]{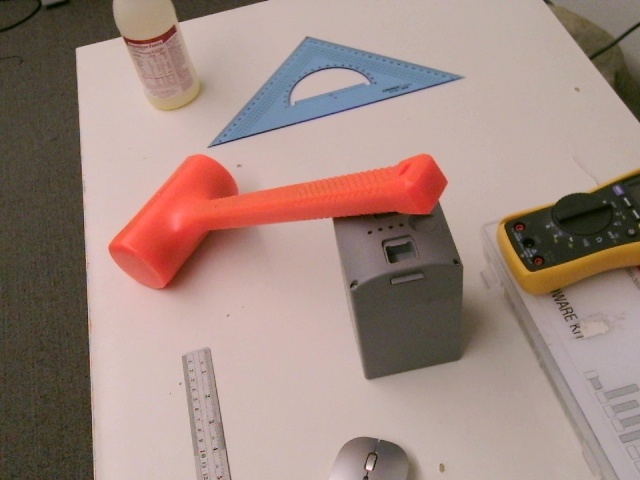}}\\    
    \fbox{\includegraphics[width=.495\linewidth]{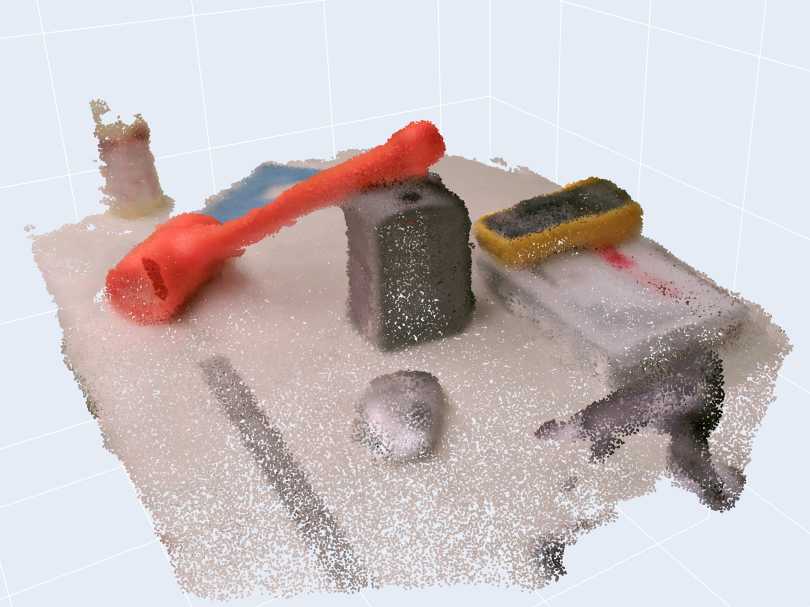}}%
    \fbox{\includegraphics[width=.495\linewidth]{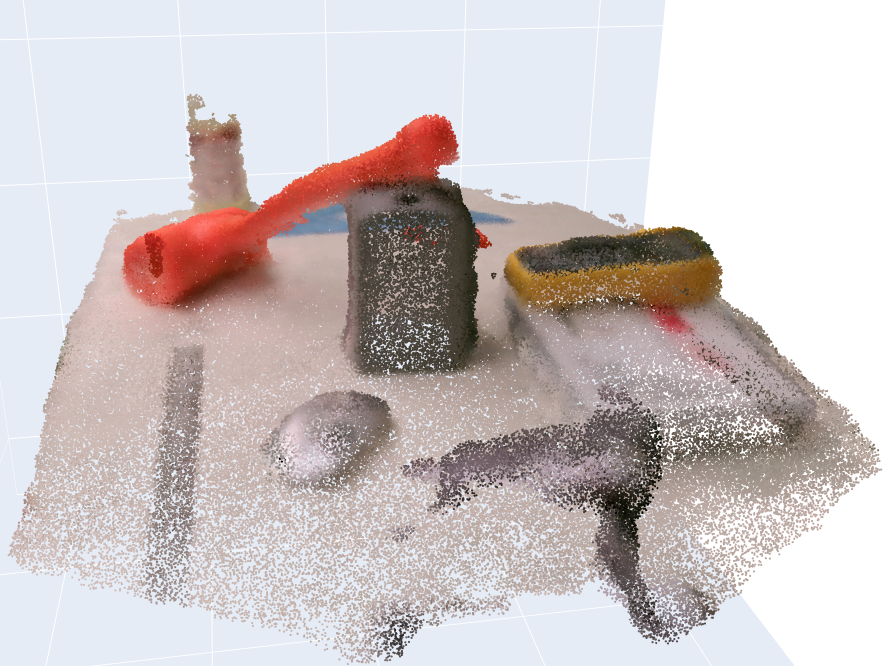}}
    \caption{Light Tabletop 8 items}
  \end{subfigure}
  \hfill % This adds space between your subfigures
  % Third subfigure
  \begin{subfigure}{.325\textwidth}
    \centering
    \fbox{\includegraphics[width=.495\linewidth]{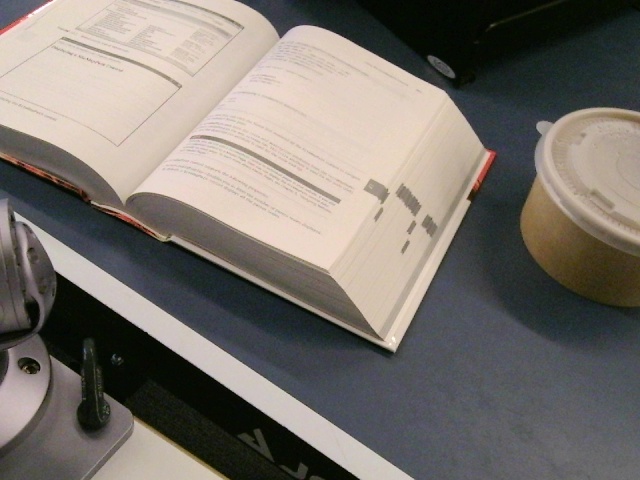}}%
    \fbox{\includegraphics[width=.495\linewidth]{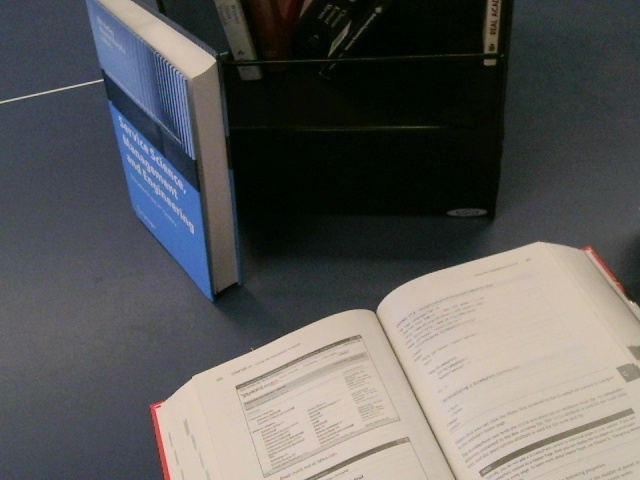}}\\
    \fbox{\includegraphics[width=.495\linewidth]{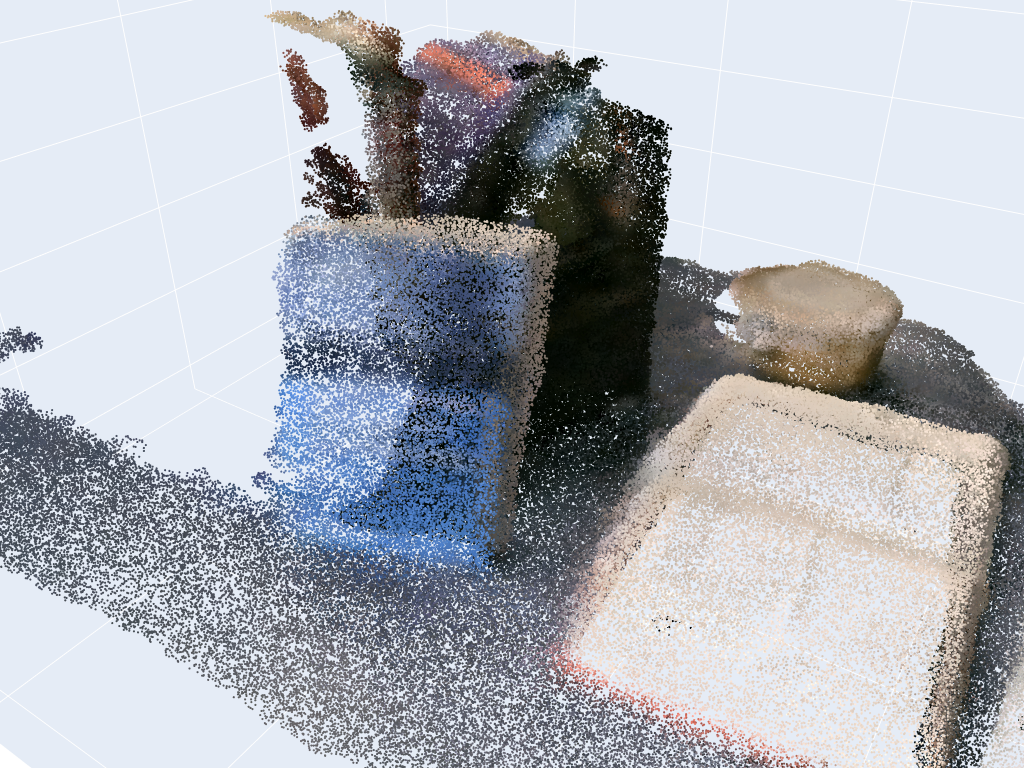}}%
    \fbox{\includegraphics[width=.495\linewidth]{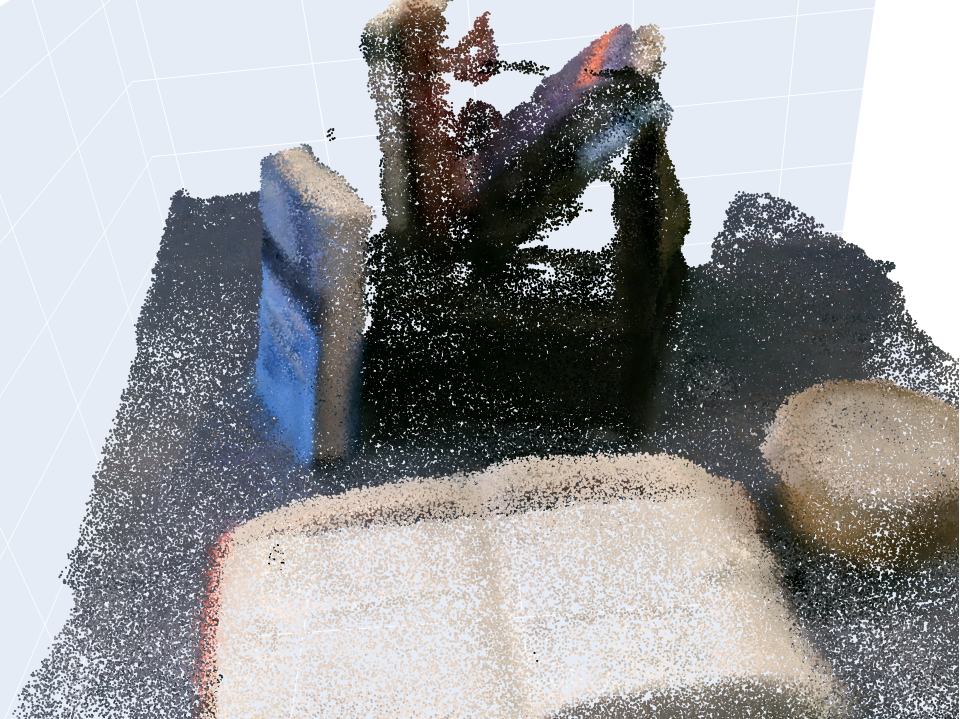}}%
    \caption{Dark Tabletop}
  \end{subfigure}
  
  \caption{\underline{Top Row}: Examples of images taken by our manipulated-mounted camera. \underline{Bottom Row}: Environment representations built with JCR. We visualize by sampling points at regions with predicted high occupancy and assign the colours predicted by the representation at these points. These representations are in the coordinate frame of the robot with physically-accurate scales.}\label{fig:constructed_envs}
\end{figure*}

\begin{figure*}[t]
\centering
\fbox{\includegraphics[width=.31\linewidth]{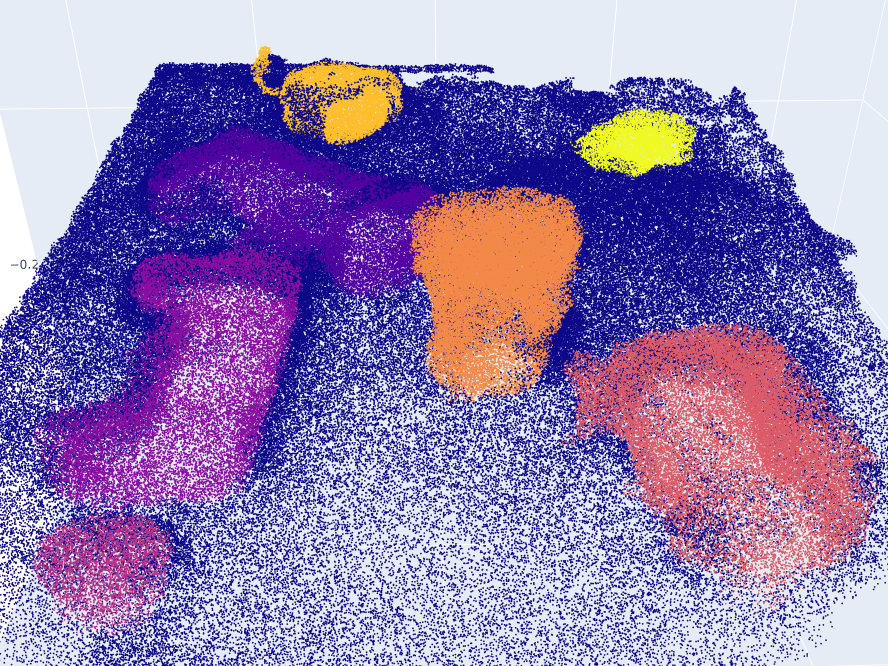}}%
\hspace{0.5em}
\fbox{\includegraphics[width=.31\linewidth]{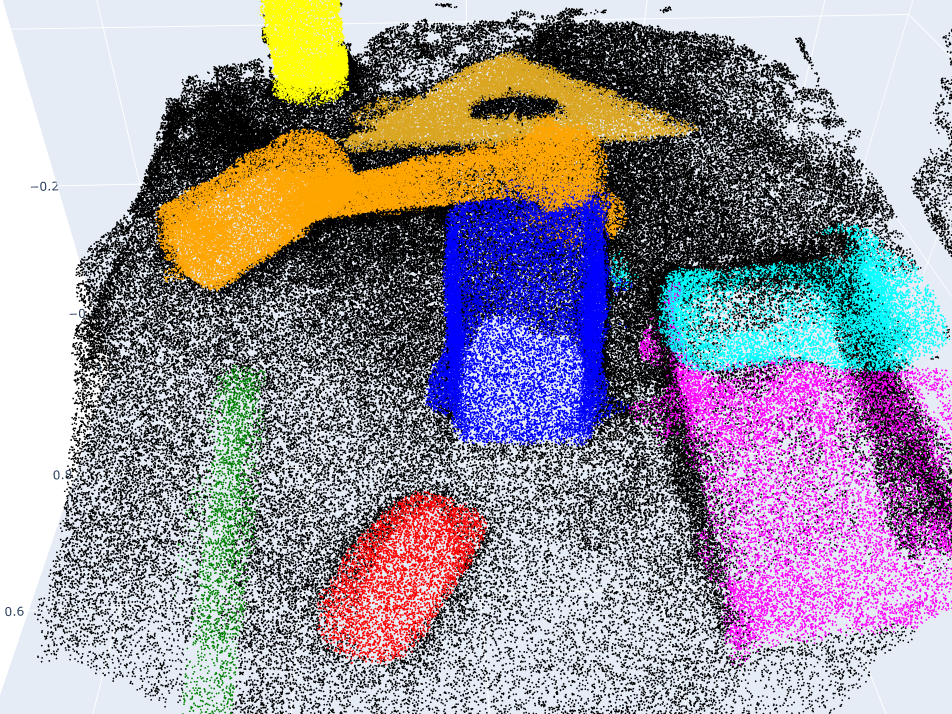}}%
\hspace{0.5em}
\fbox{\includegraphics[width=.31\linewidth]{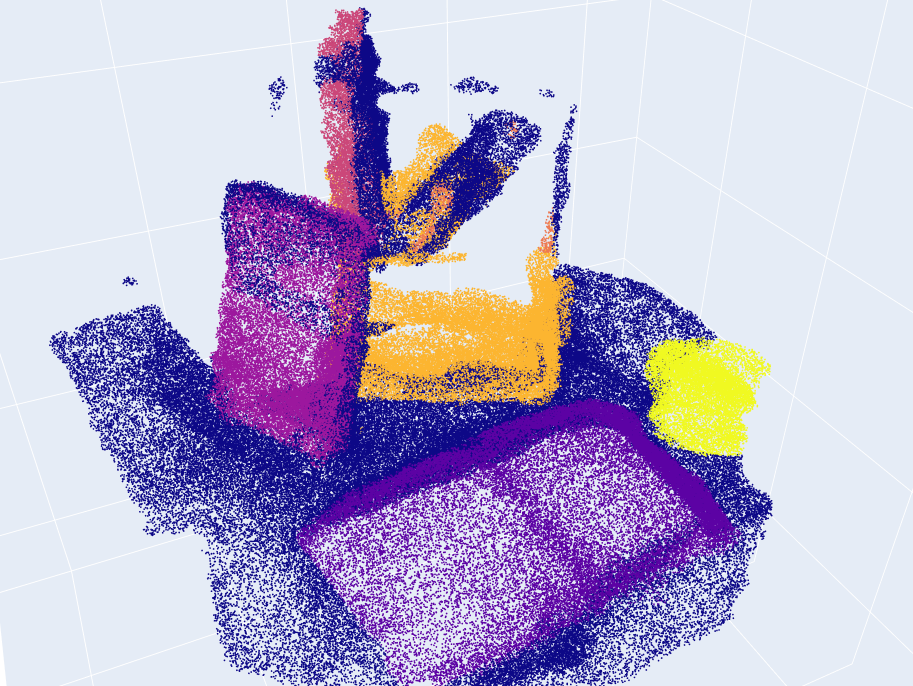}}

\caption{Segmented 3D Representations of each of the scenes (\underline{from left to right}: light tabletop 7, 8 items and dark tabletop) are visualized. We sample occupied coordinates from the occupancy representation and predict their segmentation class.}\label{fig:seg}
\end{figure*}

\subsection{Hand-Eye Calibration with JFR}
Hand-eye calibration requires the determination of relative camera poses. Historically, this has been done via artificial external markers such as checkerboards or Apriltags \cite{AprilTag}, which are highly feature-rich and easy to identify. In the absence of such markers, Structure-from-Motion (SfM) methods, such as COLMAP \cite{schoenberger2016sfm}, are typical alternative approaches to estimate relative camera poses. Here, we compare our calibration results against using COLMAP, instead of a 3D foundation model, to retrieve camera poses, with the rest of the calibration process remaining the same.

We take images in 3 different environments, two of which are table-top scenes on a light-coloured table with 2 sets of different objects with 8 and 7 items respectively, along with a scene on a dark table. We evaluate JCR with an increasing number of input images, then check whether the calibration has converged and the residual values of \Cref{eqn:cali} rearranged as 
\begin{equation}
\delta T=T_{E_{i}}^{E_{i+1}}T_{c}^{e}-T_{c}^{e}T_{P_{i}}^{P_{i+1}}(\lambda), 
\end{equation}
where lower residual values indicate a higher degree of consistency. We report the $L2$ norm of the translation term residuals $\delta_{\bb{t}}$ and the Frobenius norm of the rotation term residuals $\delta_{R}$.

% \begin{wrapfigure}{r}{0.45\linewidth} % "r" for right side, "l" for left side. Also defines the width of the wrap.
%   \centering
%   \includegraphics[width=0.98\linewidth]{colmap_overlay.png} % Change "example-image.jpg" to your image file.
%   \caption{Example of a wrapped figure.}
%   \label{fig:wrapfig}
% \end{wrapfigure}

We compare against running hand-eye calibration on camera poses estimated from COLMAP, across three different scenes. COLMAP is a widely-used SfM software, and similar to DUSt3R, estimates the relative camera poses along with the environment structure. Here, we note that running COLMAP to obtain relative camera poses is the first step in constructing NeRF \cite{mildenhall2020nerf} models, and constructing NeRF models requires successful solutions from COLMAP. We are interested in investigating the behaviour of both methods when the number of images is low: we run the methods on image sets of sizes 10, 12 and 15. The sizes of these sets of images are much lower than the image datasets used to build NeRF models, which often exceed 100 images. 

We tabulate the results in \Cref{table: res_cali}. As COLMAP relies on matching consistent hand-crafted features, many camera poses cannot be found when the number of provided images is low, resulting in divergence during calibration. On the other hand, JCR leverages foundation models to predict the correspondence and can consistently estimate relative camera poses. This results in convergent hand-eye calibration as demonstrated by the small residual sizes. As a result, JCR is more \emph{image-efficient} and allows for hand-eye calibration to be conducted even when the number of images is low.
% \begin{figure}[t]
% \centering
% \fbox{\includegraphics[width=.8\linewidth]{colmap_overlay.png}}
% \caption{Point clouds used to train JCR (in colour), produced by the upstream foundation model, is much denser than that produced by COLMAP, even after its built-in densification (overlaid in red).}\label{fig:colmap_sparse}
% \end{figure}

\begin{table}[t]
\centering

\end{table}

\subsection{Scale Recovery with JCR} 
Unlike traditional hand-eye calibration, JCR requires not only solving for the hand-eye transformation but also recovering a scale factor $\lambda$ to obtain real-world scales. Here, we run JCR on sets of 8 images and 10 images of a tabletop with a roll of tape, a box, a mug, and a toolbox (displayed in \Cref{fig:items}). We measure the heights of the objects and compare them against their respective heights in the reconstruction. The percentage height errors are given in \Cref{table:scale}, we observe that even with very few images, we can obtain sufficiently small errors. In particular, with just 10 images, the percentage errors in height for every item are at most $3.1\%$, highlighting the accuracy of the recovered scale.

\subsection{Constructing Representations with JFR}
We construct representations of the three environments to capture occupancy and colour. Neural networks with one hidden layer of size $256$ with $ReLU$ activation functions were used as the continuous representations, where each representation can be trained to convergence within 15 seconds on a standard laptop with an NVIDIA RTX 4090 GPU. We sample points at locations with high occupancy and visualize their predicted colours. We observe that JCR can construct accurate and dense representations from small sets of RGB images, without depth information.  Example images taken by the manipulator-mounted camera and visualizations of our constructed representations are provided in \Cref{fig:constructed_envs}. Additionally, we provide segmentation labels for the 2D images of each of the environments and visualize the reconstructed segmented 3D representations in \Cref{fig:seg}.

\begin{figure}[t]
  \centering
\fbox{\includegraphics[width=.8\linewidth]{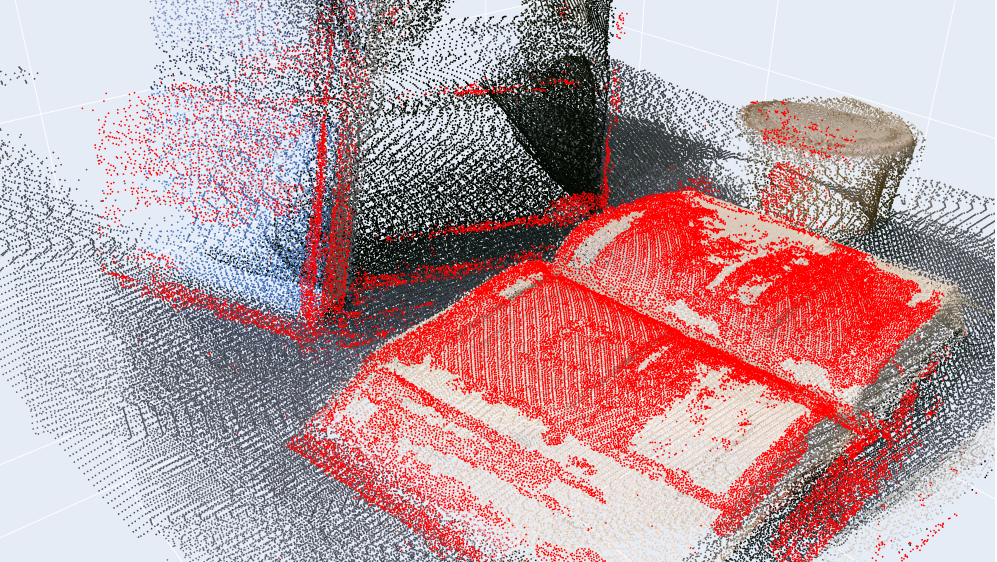}}
\caption{Point clouds produced by the foundation model for map construction in JCR are \textbf{shown in colour}. The point clouds produced by COLMAP, and after its built-in densification are overlaid and \textbf{shown in red}. COLMAP produces far sparser points, and cannot produce meaningful points on surfaces with no distinctive visual features, such as the smooth table surface.}\label{fig:colmap_sparse}
\end{figure}

As our JCR method leverages DUSt3R, which has been trained to find correspondences by predicting pointmaps, we can extract much denser representations than traditional SfM methods, which rely on visual feature-based pixel matching. In \Cref{fig:colmap_sparse}, we overlay the point clouds produced by COLMAP, followed by its built-in densification, onto points produced by the foundation model. Previous methods to create structures from multi-view images generally rely on correspondences between visual features. We observe that COLMAP cannot produce dense point clouds over smooth surfaces such as the tabletop, as the tabletop generally lacks clear features.  Instead, COLMAP can primarily identify regions that correspond to highly identifiable edges with sharp contrast, such as the text on the open book. The dense outputs of the foundation model enable us to calibrate the camera and map the environment jointly.  

% \section{Limitations}\label{sec:limitations}
% The 3D foundation model 
\section{Conclusions and Future Work}\label{sec:conclusions}
The last few years have seen the rapid boom of using large pre-trained models, or \emph{foundation models}, to facilitate a range of downstream tasks. In this paper, we advocate for the usage of foundation models to construct environment representations from a small set of images taken by a manipulator-mounted RGB camera. In particular, we propose the \emph{Joint Calibration and Representation} (JCR) method which leverages foundation models to jointly calibrate the RGB camera with respect to the robot's end-effector and construct a map. JCR enables the accurate construction of 3D representations of the environment from RBG images, in the coordinate frame of the robot without tedious \emph{a priori} calibration of the camera against external markers. We demonstrate JCR's ability to calibrate and represent the environment in an image-efficient manner, in several real-world environments. Future avenues of research include adapting JCR to dynamic environments and incorporating uncertainty information from the calibration into the constructed representations.
\bibliographystyle{ieeetr} % We choose the "plain" reference style
\bibliography{bib}
\end{document}